\documentclass{article}
\usepackage{ijcai03-6}
\usepackage{times} 
\usepackage{latexsym} 
\usepackage{theorem} 
\usepackage{amsmath} 
\usepackage{amssymb} 
\usepackage{amsfonts} 
\usepackage{graphicx} 
\usepackage{wasysym} 
\usepackage{multirow} 
\usepackage{array}

\input{epsf} 

\input xy 
\xyoption{all} 

\def\defemb#1#2{\expandafter\def\csname #1\endcsname 
  {\relax\ifmmode #2\else\hbox{$#2$}\fi}}

\defemb{cA}{{\cal A}} 
\defemb{cB}{{\cal B}} 
\defemb{cC}{{\cal C}} 
\defemb{cD}{{\cal D}} 
\defemb{cE}{{\cal E}} 
\defemb{cG}{{\cal G}} 
\defemb{cI}{{\cal I}} 
\defemb{cK}{{\cal K}} 
\defemb{cN}{{\cal N}} 
\defemb{cM}{{\cal M}} 
\defemb{cP}{{\cal P}} 
\defemb{cR}{{\cal R}} 
\defemb{cS}{{\cal S}} 
\defemb{cT}{{\cal T}} 
\defemb{cV}{{\cal V}} 
\defemb{cO}{{\cal O}}

\defemb{bF}{{\bf F}} 
\defemb{bp}{{\bf p}} 
\defemb{bU}{{\bf U}} 
\defemb{bu}{{\bf u}} 
\defemb{bV}{{\bf V}} 
\defemb{bW}{{\bf W}} 
\defemb{bw}{{\bf w}} 
\defemb{bX}{{\bf X}} 
\defemb{bx}{{\bf x}} 
\defemb{bY}{{\bf Y}} 
\defemb{by}{{\bf y}} 
\defemb{bZ}{{\bf Z}} 
\defemb{bz}{{\bf z}}

\newcommand{\qpset}{\Omega} 
\newcommand{\var}{X} 
\newcommand{\mycomment}[1]{}

\def\0{{\bf 0}} 

\def\1{{\bf 1}}

\newcommand\deff{\mathit{def}}

\newcommand{\proof}[1]{\noindent{\bf Proof:} #1 $\Box$}

\newcommand{\commentout}[1]{}

\newtheorem{theorem}{Theorem}

{\theorembodyfont{\normalfont} 
   
  } 

\begin{document} 

\title{Reasoning about soft constraints and conditional preferences:
complexity results and approximation techniques\thanks{
This research was partially funded by AFOSR, grant
F49620-01-1-0076 (Intelligent Information Systems Institute, Cornell
University) and F49620-01-1-0361 (MURI grant on Cooperative Control of
Distributed Autonomous Vehicles in Adversarial Environments),
DARPA, F30602-00-2-0530 (Controlling Computational Cost: Structure,
Phase Transitions and Randomization) and F30602-00-2-0558 (Configuring
Wireless Transmission and Decentralized Data Processing for Generic
Sensor Networks),
the Italian MIUR projects NAPOLI and COVER,
the ASI (Italian Space Agency) project ARISCOM and
Science Foundation Ireland.
%The views and conclusions contained herein are those
%of the authors and should not be interpreted as necessarily
%representing the official policies or endorsements, either expressed
%or implied, of AFOSR, DARPA, or the U.S. Government.
}} 

\author{C. Domshlak\\ Dept. of Computer Science\\
Cornell University\\
Ithaca, NY, USA \\
{dcarmel@cs.cornell.edu}
\And 
F. Rossi \and K. B. Venable \\ 
Dept. of Mathematics\\
University of Padova\\
Padova, Italy\\
{\{frossi,kvenable\}@math.unipd.it} 
\And T. Walsh \\ 
Cork Constraint Computation Centre\\
University College Cork\\ 
Cork, Ireland\\
{tw@4c.ucc.ie}} 

%Content areas: 
%constraint satisfaction, knowledge representation.}

\maketitle 

\begin{abstract} 
%\begin{quote}   
  Many real life optimization problems contain
  both hard and soft constraints, as well as qualitative conditional
  preferences.  However, there is no single formalism to specify
  all three kinds of information.  We therefore propose a framework,
  based on both CP-nets and soft constraints, that handles both hard and 
  soft constraints as well as conditional preferences efficiently and uniformly. 
  We study the complexity of testing the consistency of preference statements, 
  and show how soft constraints can faithfully approximate the semantics of 
conditional
  preference statements whilst improving the computational complexity.
%\end{quote} 
\end{abstract}

\section{Introduction and Motivation} 

% carmel : I moved this paragraph forward
Representing and reasoning about preferences is an area of increasing interest
in theoretical and applied AI.  In many real life problems, we have both hard
and soft constraints, as well as qualitative conditional preferences.  For
example, in a product configuration problem, the producer may have hard and
soft constraints, while the user has a set of conditional preferences.
Until now, there has been no single formalism which allows all
these different kinds of information to be specified
efficiently and reasoned with effectively.
For example, soft constraint solvers \cite{jacm,schiex-ijcai95} 
are most suited for reasoning about the hard and soft
% qualitative
constraints, while CP-nets~\cite{BBHP.UAI99}
are most suited for representing 
% are thought for
qualitative conditional preference statements.  In this paper, we exploit a
connection between these two approaches, and define a framework based on both
CP-nets and soft constraints which can efficiently handle both constraints and
preferences.

{Soft constraints} \cite{jacm,schiex-ijcai95} 
are one of the main methods for dealing with
preferences in constraint optimization. Each assignment to the variables of
a constraint is annotated with a level of its desirability, and the
desirability of a complete assignment is computed by
a combination operator applied to the ``local'' preference values. Whilst
soft constraints are very expressive, and have a 
powerful computational machinery, they
are not good at modeling and solving the sort of 
conditional preference statements that occur in the real world.  
Moreover, soft constraints are based
on {\em quantitative}
measures of preference, which tends to make preference elicitation
more difficult.

Qualitative user preferences have been widely studied in
decision-theoretic AI~\cite{Doyle.survey.99}.  Of particular interest 
are CP-nets~\cite{BBHP.UAI99}. These 
model statements of qualitative and conditional preference such as ``I prefer
a red dress to a yellow dress", or ``If the car is convertible, I prefer a soft
top to a hard top''. These are interpreted under the 
{\em ceteris paribus} (that is,
``all else being equal'') assumption. 
Preference elicitation in such a framework is intuitive, independent
of the problem constraints, and suitable %even 
for naive users. 
However, the Achilles
heel of CP-nets and other sophisticated qualitative preference
%representation 
models~\cite{Lang:kr02} is the complexity of reasoning
with them~\cite{Domshlak:Brafman:kr02,BBDHP.journal}.
%,Domshlak:PhD}. 

Motivated by a product configuration
application~\cite{Sabin:Weigel:ieee98}, we have developed a framework 
to reason simultaneously about qualitative conditional preference statements 
and hard and soft constraints. In product configuration,
the producer has %both 
hard (e.g., component compatibility) and soft 
(e.g., supply time)
constraints, while the customer has preferences over the product features.
We first investigate the complexity of reasoning
about qualitative preference statements, addressing in particular %the issue of
preferential consistency. To tackle the complexity of 
preference reasoning, we then introduce two
approximation schemes based on soft constraints. 

To the best of our knowledge, this work
provides the first connection between the CP-nets and soft constraints
machinery. 
% carmel : I moved the following sentence from the end to here
In addition, for product configuration problems or any problem with
both hard and soft quantitative constraints as well as 
qualitative conditional preferences, this framework lets us treat 
the three kinds of information in a unifying
environment.
Finally, we 
compare the two approximations in terms of both
expressivity and complexity. 
% and compared to previous approximations of CP-nets.

\section{Formalisms for Describing Preferences} 

\subsection{Soft constraints} 

There are many formalisms for describing {\em soft constraints}.  We use the 
c-semi-ring formalism \cite{jacm}, which is 
equivalent to the valued-CSP formalism when total orders are used
\cite{merge}, 
as this generalizes many of the others.
%(\cite{schiex,fuzzy1,partial-ai}).  
In brief, a soft constraint 
associates each instantiation of its variables with a 
value from a partially ordered set.  We also supply operations for combining 
($\times$) and comparing (+) values.  A semi-ring is a tuple $\langle 
A,+,\times,\0,\1 \rangle$ such that: $A$ is a set and $\0, \1 \in A$; $+$ is 
commutative, associative and $\0$ is its unit element; $\times$ is associative, 
distributes over $+$, $\1$ is its unit element and $\0$ is its absorbing 
element.  A {\em c-semi-ring} is a semi-ring $\langle A,+,\times,\0,\1 \rangle$ 
in which $+$ is idempotent, $\1$ is its absorbing element and $\times$ is 
commutative. 

Let us consider the relation $\leq$ over $A$ such that $a \leq b$ iff $a+b = 
b$. Then $\leq$ is a partial order, $+$ and $\times$ are monotone on $\leq$, 
$\0$ is its minimum and $\1$ its maximum, $\langle A,\leq \rangle$ is a 
complete lattice and, for all $a, b \in A$, $a+b = lub(a,b)$.  Moreover, if 
$\times$ is idempotent: $+$ distributes over $\times$; $\langle A,\leq \rangle$ 
is a complete distributive lattice and $\times$ its glb. 
% carmel: line break removed 
% -------------------------- 
% 
Informally, the relation $\leq$ 
compares semi-ring values and constraints. 
When $a \leq b$, we say that {\em b is better 
than a}. 
% carmel: line break removed 
% -------------------------- 
% 
Given a semi-ring $S = \langle A,+,\times,\0,\1 \rangle$, a finite 
set $D$ (variable domains) and an ordered set of 
variables $V$, a {\em constraint} is a pair $\langle 
\deff, con \rangle$ where $con \subseteq V$ and $\deff: D^{|con|} 
\rightarrow A$. A constraint specifies a set of 
variables, and assigns to each tuple of values 
of these variables an element of the semi-ring. 

% carmel: two paragraphs below rephrased (saving space) and joined 
% ---------------------------------------------------------------- 
% A {\em soft constraint satisfaction problem} (SCSP) is just a set of soft 
% constraints.  A classical constraint satisfaction problem is an SCSP with the 
% c-semi-ring: $S_{CSP} = \langle \{false, true\},\vee, \wedge, false, true 
% \rangle$. Fuzzy CSPs \cite{schiex} can be modelled in the SCSP framework by 
% choosing the c-semi-ring $S_{FCSP} = \langle [0,1], max, min, 0, 1 \rangle$. 
% Many other types of soft constraints (probabilistic, weighted, \ldots) can be 
% modeled by using suitable semi-rings ($S_{prob} = \langle [0,1], max, \times, 
% 0, 1 \rangle$, $S_{weight} = \langle \mathcal{R}, min, +, 0, +\infty \rangle$, 

% \ldots). 
% 
% A solution to an SCSP is an assignment of all its variables to elements in 
% their domains. The preference value associated to a solution is obtained by 
% multiplying the preference values of the projections of the solution over each 

% constraint.  A solution is better than another if its preference value is 
% higher in the partial order of the semi-ring. 
A {\em soft constraint satisfaction problem} (SCSP) is given by a set of soft 
constraints. For example, a classical CSP is an SCSP with the c-semi-ring 
$S_{CSP} = \langle \{false, true\},\vee, \wedge, false, true \rangle$, a fuzzy 
CSP~\cite{schiex} is an SCSP with the c-semi-ring $S_{FCSP} = \langle [0,1], 
max, min, 0, 1 \rangle$, and probabilistic and weighted CSPs are SCSPs with the 
c-semi-rings $S_{prob} = \langle [0,1], max, \times, 0, 1 \rangle$ and 
$S_{weight} = \langle \mathcal{R}, min, +, 0, +\infty \rangle$, respectively. 
A solution to an SCSP is a complete assignment to its variables. The 
preference value associated with a solution is obtained by multiplying the 
preference values of the projections of the solution to each constraint. One
solution is better than another if its preference value is higher in the 
partial order. % of the semi-ring. 
Finding an optimal solution for an SCSP is an {\sc np}-complete problem. On 
the other hand, given two solutions, checking whether one is 
% carmel: replaced 
% ---------------- 
% more preferred 
preferable 
% 
%to another 
is easy: we compute the semi-ring values of the two 
solutions and compare the resulting values.

\subsection{CP-nets} 
\label{cpback} 

Soft constraints are the main tool for representing and reasoning about 
preferences in constraint satisfaction problems. However, they require 
the choice of a semi-ring value for each variable assignment in each constraint.
They are therefore a {\em quantitative} method for 
expressing preferences.  In many applications, it is more natural for users 
to express  preferences via generic qualitative (usually 
partial) preference relations over variable assignments. 
For example, 
% carmel : replaced 
% ----------------- 
% we would rather say 
it is often more intuitive for the user to say 
``I prefer red wine to white wine'', rather 
than ``Red wine has preference 0.7 and white wine has preference 0.4''. 
% carmel : removed (clear from the context). 
% (with the assumption that a higher preference value is better). 
% 
% carmel : replaced 
% Of course the first sentence has less information, but it is more natural for 
% users, and does not require a careful preference value selection to maintain 
% consistency over the whole problem. 
Of course, the former statement provides less information, but it does not 
require careful selection of preference values to maintain consistency. 
Moreover, soft constraints do not naturally represent conditional preferences, 
as in ``If they serve meat, then I prefer red wine to white wine''.  It is easy 
to see that both qualitative statements and conditions are essential 
ingredients in many applications.

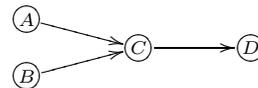
\begin{figure}[t] 
\begin{center} 
  %\ \setlength{\epsfxsize}{1.8in} 
  {\scriptsize 
  \mbox{ 
    \entrymodifiers={+[o][F-]} 
    \def\objectsizestyle{\scriptstyle} 
    \xymatrix @R=0pt@C=30pt{ 
      A \ar[dr]\\ 
      *+{} & C \ar[r] & D\\ 
      B \ar[ur] 
      } 
    } 
  } 
\end{center} 
\caption{The CP-net graph for the example.} 
%\vspace{-0.5cm} 
\label{cpnet} 
\end{figure}

% carmel : replaced 
% CP-nets were introduced in~\cite{BBHP.UAI99} as a tool 
CP-nets~\cite{BBHP.UAI99} are a graphical model 
for compactly representing conditional and qualitative preference relations.
They exploit conditional preferential independence by structuring a 
user's preferences under the {\em ceteris paribus} assumption. Informally, 
CP-nets are sets of {\em conditional ceteris paribus (CP)\/} preference 
statements. For instance, the statement {\em "I prefer red wine to white 
  wine if meat is served."} asserts 
that, given two meals that differ {\em only} in the kind of wine served {\em 
  and} both containing meat, the meal with a red wine is preferable to the meal 
with a white wine.  Many philosophers 
(see~\cite{Hansson:01} 
% (see~\cite{Hansson:96,Hansson:01} 
for an overview) and AI 
researchers~\cite{DoyleWellman.AAAI95}, 
% ~\cite{DoyleShohamWellman.91,DoyleWellman.AAAI95}, 
have argued that
most of our preferences are of this type. 

%carmel-f : rephrased
% CP-nets bear some similarity to Bayesian networks. Both utilize
CP-nets bear some similarity to Bayesian networks, as both utilize
directed acyclic graphs where each node stands for a domain variable,
and assume a set of features $\bF = \{\var_1,\ldots,\var_n\}$ with
finite, discrete domains $\cD(\var_1),\ldots,\cD(\var_n)$ (these play
the same role as variables in soft constraints).  During preference
elicitation, for each feature $\var_i$, the user is asked to specify a
set of {\em parent} features $Pa(\var_i)$ that can affect her
preferences over the values of $\var_i$.  This information is used to
create the graph of the CP-net in which each node $\var_i$ has
$Pa(\var_i)$ as its immediate predecessors.  Given this structural
information, the user is asked to explicitly specify her preference
over the values of $\var_i$ for
%carmel-f : replaced
% each assignment
{\em each complete assignment} on $Pa(\var_i)$, and this
preference is assumed to take the form of total~\cite{BBHP.UAI99} or
partial~\cite{BBDHP.journal} order over $\cD(X)$.  These conditional
preferences over the values of $\var_i$ are
%captured by a {\em conditional preference table} which is 
annotated with the node $\var_i$ in the CP-net.  For example, consider
a CP-net with the graph given in Figure~\ref{cpnet}, and with the
preference statements as follows: $a \succ \overline{a}$, $b \succ
\overline{b}$, $(a \wedge b) \vee (\overline{a} \wedge \overline{b}) :
c \succ \overline{c}$, $(a \wedge \overline{b}) \vee (\overline{a}
\wedge b) : \overline{c} \succ c$, $c: d \succ \overline{d}$,
$\overline{c}: \overline{d} \succ d$.  Here, statement $a \succ
\overline{a}$ represents the unconditional preference of the user for
$A=a$ over $A=\overline{a}$, while statement $c: d \succ \overline{d}$
represents that the user prefers $D=d$ to $D=\overline{d}$, given that
$C=c$.

Several types of queries can be asked about CP-nets.  First, given a 
CP-net $N$, one might be interested in finding an optimal assignment to the 
features of $N$. For acyclic CP-nets, such a query is answerable 
in linear time~\cite{BBHP.UAI99}. Second, given a CP-net $N$ and a pair of 
complete assignments $\alpha$ and $\beta$, one 
might be interested in determining 
whether %$N$ implies 
$\alpha \succ \beta$, i.e.\ $\alpha$ is preferred to 
$\beta$.  Unfortunately, this query  is 
{\sc np}-hard even for acyclic CP-nets~\cite{Domshlak:Brafman:kr02} though
some tractable special cases do exist.

\section{Consistency and Satisfiability} 
Given a set of preference statements $\qpset$ extracted from a user,
we might be interested in testing {\em consistency} of the induced
preference relation.  In general, there is no single notion of
preferential consistency~\cite{Hansson:01}.  In~\cite{BBHP.UAI99}, a
CP-net $N$ was considered consistent iff the partial ordering $\succ$
induced by $N$ %over the outcomes 
is {\em asymmetric}, i.e.\ there
exist at least one total ordering of the outcomes consistent with
$\succ$.  However, in many situations, we can ignore cycles in the
preference relation, as long as these do not prevent %the user making 
a rational choice, i.e.\ there exist an outcome %$\alpha$ 
that is not
dominated by any other outcome. % with respect to $\succ$.  
In what
follows, we refer to this as {\em satisfiability}\footnote{In
  preference logic~\cite{Hansson:01}, these notions of ``consistency
  as satisfiability'' and ``consistency as asymmetry'' correspond to
  the notions of {\em eligibility} and {\em restrictable eligibility},
  respectively. However, we will use the former terms as they seem
  more intuitive.}.  It is easy to see that satisfiability is strictly
weaker than asymmetry, and that asymmetry implies satisfiability.
%carmel-f : the first sentence has been joined to the above paragraph
%           and an additional clarifying sentence has been added
We will consider two cases: When the set $\Omega$ of preference statements
induces a CP-net and, more generally, when preferences can take any
form (and may not induce a CP-net). 
%The latter case corresponds to a
%situation in which there exists a variable $X$, such that the
%conditionals (i.e.\ contexts) of the preference statements over the
%value of $X$ in $\Omega$ are not mutually exclusive.

When $\qpset$ defines an acyclic CP-net, the partial 
order induced by $\qpset$ is asymmetric~\cite{BBHP.UAI99}.  However, for 
cyclic CP-nets, asymmetry is not guaranteed. 
In the more general case, we are 
given a set $\qpset$ of conditional preference statements without
any guarantee that they define a CP-net. Let the 
{\em dependence graph} of $\qpset$ be defined similarly to the graphs of 
CP-nets: the nodes stand for problem features, and a directed arc 
goes from $X_i$ to $X_j$ iff $\qpset$ contains a 
statement expressing preference on 
the values of $X_j$ conditioned on the value of $X_i$. 
For example, the 
set $\qpset = \{a : b \succ \overline{b}, a \wedge c : \overline{b} \succ b\}$ 
does not induce a CP-net
%carmel-f : added explanation in brackets
(the two conditionals are not mutually exclusive),
and the preference relation induced by $\qpset$ is not asymmetric, 
despite the fact that the dependence graph of $\qpset$ is acyclic.
% carmel -- if we'll have some space, the remarked sentences below 
%           will better be restored. 
% ---------------------------------------------------------------- 
%Assuming that the {\em true} 
%preferences of the user are asymmetric, given such a set of statements, 
% one can 
%try to continue questioning the user, trying to eliminate the in-asymmetry 
% that has been detected. 
% However, (even ignoring the fact that detecting asymmetry can be 
%hard in general) in online configuration tasks that are of our interest, long 
%interactions with the user should be prevented as much as possible. 

Note that while asymmetry implies satisfiability, the reverse does not hold 
in general. For example, the set $\qpset$ above is not asymmetric, but it is 
satisfiable (the assignment $a\overline{c}b$ is undominated). Given such a 
satisfiable set of statements, we can prompt the user with one of the 
undominated assignments without further refinement of its preference relation. 
Theorem~\ref{t:hard} shows that, in general, determining satisfiability of a set 
of statements is {\sc np}-complete. On the other hand, even for CP-nets, 
determining asymmetry is not known to be in {\sc 
  np}~\cite{Domshlak:Brafman:kr02}. 

\begin{theorem} 
\label{t:hard} 
{\sc satisfiability} of a set of conditional preference statements $\qpset$ is 
{\sc np}-complete. 
\end{theorem} 

\proof{ Membership in {\sc np} is straightforward, as an assignment is a 
  polynomial-size witness that can be checked for non-dominance in time linear 
  in the size of $\qpset$.  To show hardness, we reduce {\sc 3-sat} to our 
  problem: Given a 3-cnf formula $F$, for each clause $(x \vee y \vee z) 
  \in F$ we construct the conditional preference statement: $\overline{x} 
  \wedge \overline{y} : z \succ \overline{z}$.  \mycomment{Consider any model 
    (satisfying assignment) of the original 3-cnf formula.  At least one of 
    $x$, $y$ and $z$ will be set true. If $x$ or $y$ are true, then the 
    condition of the constructed conditional preference constraint is not 
    satisfied and we can ignore it. If $x$ and $y$ are both false, then $z$ 
    must be true.  However, setting $z$ to true satisfies the conditional 
    preference ordering as this is the most preferred value. Hence, any model 
    of the original 3-cnf formula is an optimal assignment of the constructed 
    CP-net. The argument reverses and any optimal assignment is also a model. 
    The CP-net is therefore satisfiable iff the original 3-cnf problem is 
    satisfiable.}  This set of conditional preferences is satisfiable iff the 
  original original formula $F$ is satisfiable.  } 
   
While testing satisfiability is hard in general, Theorem~\ref{t:acyclic} 
presents a wide class of statement sets that can be tested for 
satisfiability in polynomial time. 

\begin{theorem} 
  \label{t:acyclic} 
  A set of conditional preference statements $\qpset$, whose dependency graph 
  is acyclic and has bounded node in-degree can be tested for 
  satisfiability in polynomial time. 
\end{theorem} 

\proof{ The proof is constructive, and the algorithm is as follows: First, for
  each feature $X \in \bV$, we construct a table $T_{X}$ with an entry for each
  assignment $\pi \in \cD(Pa(X))$, where each entry $T_{X}[\pi]$ contains all
  the values of $X$ that are not dominated given $\qpset$ and $\pi$.
  Subsequently, we remove all the empty entries. For example, let $A$, $B$ and
  $C$ be a set of boolean problem features, and let $\qpset = \{c \succ
  \overline{c}, a : b \succ \overline{b}, a \wedge c : \overline{b} \succ b\}$.
  The corresponding table will be as follows:
  {\scriptsize
  \begin{center} 
    \begin{tabular}{|c|c|c|} 
      \hline 
        Feature & $\pi$ & Values\\ 
        \hline 
        \hline 
        $T_A$ & $\emptyset$ & $\{a,\overline{a}\}$\\ 
        \hline 
        \hline 
        $T_C$ & $\emptyset$ & $\{c\}$\\ 
        \hline 
        \hline 
        $T_B$ & $a \wedge \overline{c}$ & $\{b\}$\\ 
        \hline 
        & $\overline{a} \wedge \overline{c}$ & $\{b,\overline{b}\}$\\ 
        \hline 
        & $\overline{a} \wedge c$ & $\{b,\overline{b}\}$\\ 
        \hline 
    \end{tabular} 
  \end{center}
  }
  Observe that the entry $T_B[a \wedge c]$ has been removed, since, given $a 
  \wedge c$, $b$ and $\overline{b}$ are dominated according to the statements 
  $a \wedge c : \overline{b} > b$ and $a : b > \overline{b}$, respectively. 
  Since the in-degree of each node $X$ in the dependence graph of $\qpset$ is 
  bounded by a constant $k$ (i.e.\ $|Pa(X)| \leq k$), these tables take space 
  and can be constructed in time $O(n2^k)$.  Given such tables for all the 
  features in $\bV$, we traverse the dependence graph of $\qpset$ in a 
  topological order of its nodes, and for each node $X$ being processed we 
  remove all the entries in $T_X$ that are not ``supported'' by (already 
  processed) $Pa(X)$: An entry $T_X[\pi]$ is not supported by $Pa(X)$ if there 
  exists a feature $Y \in Pa(X)$ such that the value provided by $\pi$ to $Y$ 
  appears in no entry of $T_Y$.  For instance, in our example, the rows 
  corresponding to $a \wedge \overline{c}$ and $\overline{a} \wedge 
  \overline{c}$ will be removed, since $\overline{c}$ does not appear in the 
  (already processed) table of $C$. Now, if the processing of a feature $X$ 
  results in $T_X = \emptyset$, then $\qpset$ is not satisfiable.  Otherwise, 
  any assignment to $\bV$ consistent with the processed tables will be 
  non-dominated with respect to $\qpset$.} 

Note that, for sets of preference statements with cyclic dependence graphs,
{\sc satisfiability} remains hard even if the in-degree of each node is bounded
by $k \geq 6$, since {\sc 3-sat} remains hard even if each variable
participates in at most three clauses of the formula
%~\cite{Garey:Johnsson} (see
the proof of Theorem~\ref{t:hard}).  However, when at most one condition is
allowed in each preference statement, and the features are boolean, then {\sc
  satisfiability} can be reduced to {\sc 2-sat}, and thus tested in polynomial
time.  Further study of additional tractable cases is clearly of both
theoretical and practical interest.

\section{Approximating CP-nets with Soft Constraints} 

In addition to testing consistency and determining preferentially optimal
outcomes, we can be interested in the {\em preferential comparison} of
two outcomes.  
%Comparison is essential in
%preference-based optimization in face of some hard constraints on the
%variables, and in sorting a predefined set of outcomes (e.g., the content of a
%database relation).  
Unfortunately, determining dominance between a pair of
outcomes with respect to a set of qualitative preferential statements under
the {\em ceteris paribus} assumption is {\sc pspace}-complete in
general~\cite{Lang:kr02}, and is {\sc np}-hard even for acyclic
CP-nets~\cite{Domshlak:Brafman:kr02}. However, given a set $\qpset$ of
preference statements, instead of using a preference relation $\succ$ induced by
$\qpset$, one can use an approximation $\gg$ of $\succ$, achieving tractability
%of comparison 
while sacrificing precision to some degree.
Clearly, different approximations $\gg$ of $\succ$ are not equally good, as
they can be characterized by the precision with respect to $\succ$, time
complexity of generating $\gg$, and time complexity of comparing outcomes with
respect to $\gg$. In addition, it is vital that
$\gg$ faithfully {extends} $\succ$ (i.e.\ $\alpha \succ \beta$ should entail
$\alpha \gg \beta$). We call this {\em information
preserving}.
% carmel1 : rephrased
Another desirable property of approximations is that of
preserving the {\em ceteris paribus} property
(we call this the {\em cp-condition} for short).
% It is also important that approximations respect the Ceteris
% Paribus condition, that guarantees the dependence of a feature on a set of
% parent features ``all else being equal'', that is, when the remaining features
% are assigned the same values.

For acyclic CP-nets, two approximations that are
information preserving have been introduced, % in the literature, 
both comparing outcomes in time linear in the number of features.  The
first is based on the relative position of the features in the CP-net
graph~\cite{BBDHP.journal}.
%,Domshlak:PhD}. 
This approximation does not require any preprocessing of the CP-net.  However,
it is problematic when there are hard constraints.  
The second, based on UCP-nets~\cite{Boutilier:Bacchus:Brafman:uai01}, 
can be used as a quantitative approximation of acyclic CP-nets. 
UCP-nets resemble weighted CSPs, and thus they can be used in 
constraint optimization using the
soft constraints machinery. However, generating UCP-nets is 
exponential in the
size of CP-net node's Markov family\footnote{{\em Markov family} 
of a node $X$ contains $X$, its parents and children, and 
the parents of its children.},
and thus in the CP-net node out-degree.
%carmel-f : rephrased as the scenario is not really similar
%A similar scenario 
An additional related work
is described in \cite{md-aaai2002},
%carmel-f : rephrased
where a numerical value function is constructed using graph-theoretic
techniques by examining the graph of the preference relation induced by
a set of preference statements.  Note that this framework is also
computationally hard, except for some special cases.

\begin{figure}[t] 
\begin{center}
  %\ \setlength{\epsfxsize}{3.3in}
  \ \setlength{\epsfxsize}{2.8in}
\epsfbox{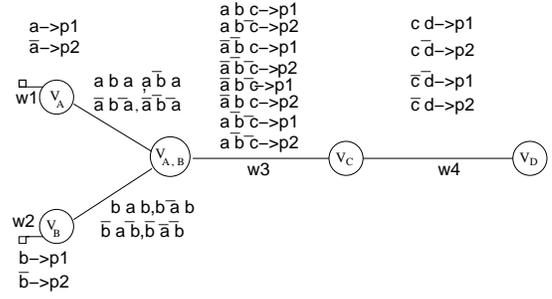} 
\end{center} 
%\vspace{-0.5cm} 
\caption{An SC-net.}\label{scnet} 
%\vspace{-0.3cm} 
\end{figure}

Here we study approximating CP-nets via soft constraints (SCSPs).
This allows us to use the rich machinery underlying SCSPs to answer comparison
queries in linear time.  Moreover, this provides us a
uniform framework to combine user preferences %over the outcomes 
with both hard and soft constraints.
Given an acyclic CP-net, we construct a corresponding SCSP in two steps. First,
we build a constraint graph, which we call {\em SC-net}.  Second, we compute
the preferences and weights for the constraints in the SC-net, and this
computation depends on the actual semi-ring framework being used.  Here we
present and discuss two alternative semi-ring frameworks, based on {\em min+}
and {\em SLO} (Soft constraint Lexicographic Ordering) semi-rings, respectively. In both cases, our computation of
preferences and weights ensures information preserving and satisfies the
cp-condition.  We illustrate the construction of the SCSP using the example in
Figure~\ref{scnet}, which continues our running example from Figure
\ref{cpnet}.

Given a CP-net $N$, the corresponding SC-net $N_c$ has two types of 
nodes: First, each feature $X \in N$ is represented in $N_c$ by a node $V_X$ 
that stands for a SCSP variable with $\cD(V_X) = \cD(X)$. Second, for each 
feature $X \in N$, such that $|Pa(X)| \geq 2$, we have a node $V_{Pa(X)} \in 
N_c$, with $\cD(V_{Pa(X)}) = \Pi_{Y \in Pa(X)} \cD(Y)$. 
% carmel -- are you sure that people will understand that \Pi stand for 
%           crossproduct? 
Edges in $N_c$ correspond to hard and soft constraints, where the latter are 
annotated with weights. Each node $V_X$ corresponding to an 
``independent feature'' $X\in N$ has an incoming (source-less) soft constraint 
edge (e.g., see $V_A$ and $V_B$). For each node $V_X$ corresponding to 
a "single-parent" feature $X\in N$ with $Pa(X) = \{Y\}$, we have a soft 
constraint edge between $X$ and $Y$ (e.g., see 
$V_D$). Finally, for each node $V_X$ such that 
$|Pa(X)| \geq 2$, we have (i) hard constraint edges between $V_{Pa(X)}$ and 
each $Y \in Pa(X)$ to ensure consistency (e.g., the edges between $V_{A,B}$ and 
both $V_A$ and $V_B$), and (ii) a soft constraint edge between $V_{Pa(X)}$ and 
$V_X$ (e.g., the edge between $V_{A,B}$ and $V_C$).

To assign preferences to variable assignments in each soft constraint, 
each soft constraint $c$ (between $V_{Pa(X)}$ and $V_X$) 
is associated with two items: 
$w_c$, a real number which can be interpreted as a weight
(will be defined in the next section), and 
$P_c = \{p_1, ..., p_{|\cD(V_X)|}\}$, a set of reals
which can be interepreted as ``quantitative levels of preference''.
We will see in the next section how to generate the 
preference for each assignment to the variables of $c$,
depending on the chosen semiring.
In any case, each preference will be obtained by combining 
(via multiplication over naturals) 
the weight of the constraint $w_c$ and one of the elements of $P_c$.

\subsection{Weighted soft constraints} 

The weighted SCSP is based on 
the {\em min+} semi-ring $S_{WCSP}=\langle R_+,
min, +, +\infty , 0\rangle$.  We assign preferences 
using real positive numbers 
(or penalties) and prefer assignments with smaller 
total penalty (i.e.\ the sum
of all local penalties). In a soft constraint $c$ on $V_{Pa(X)}$ 
and $V_X$, 
there are $|\cD(V_X)|$ penalties.  
Without loss of generality, we assume they range 
between $0$ and $|\cD(V_X)|-1$, that is, $p_1=0, ..., p_{\cD(V_X)|} = 
|\cD(V_X)|-1$.
In our example, since all
variables are binary, there are only two penalties 
i.e., $p1=0$ and $p2=1$, in all the constraints.

\begin{figure}[b] 
\begin{center} 
  {\footnotesize 
\fbox{ 
\begin{minipage}{3.2in} 
\begin{tabbing} 
xx\=xx\=xx\=xx\=xx\=xx\= \kill 
{${\mathsf{Input:}}$} Acyclic CP-net $N$\\ 
1. Construct the SC-net $N_c$ without weights.\\ 
2. Order variables of $N$ in a reverse topological ordering.\\ 
3. {\bf foreach} $X \in N$ {\bf do}\\ 
4. \>\> {\bf if} $X$ has no successors in $N$ {\bf then}\\ 
5. \>\>\> $w(V_X) = 1$\\ 
6. \>\> {\bf else}\\ 
7. \>\>\> $w(V_X) = \sum_{Y\;{\mathrm{s.t.}}\;X\in 
Pa(Y)}w(V_Y)\cdot|\cD(V_Y)|$\\ 
8. {\bf return} $N_c$ 
\end{tabbing} 
\end{minipage} 
}} 
\end{center} 
\caption{Algorithm for weight computation.}\label{impdom} 
%\vspace{-0.4cm} 
%the {\em ceteris paribus} update of SC-nets.} 
\end{figure}

To ensure the cp-condition, similar to~\cite{Boutilier:Bacchus:Brafman:uai01}, 
we need to ensure that each variable
dominates its children.
%In fact, this condition is sufficient for the ceteris
%paribus property, and easier to build.
We therefore set the minimum penalty on a variable to be greater than the sum
of the maximum penalties of the children.  In Figure \ref{impdom} we show the
pseudocode for the algorithm to compute the weights.  In this code, $w(V_X)$
represents the weight of the soft constraint $c$ between $V_{Pa(X)}$ and $V_X$.

Considering our example, let $\{D,C,B,A\}$ be the reverse topological ordering
obtained in line 2. Therefore, the first soft constraint to be processed is the
one between $V_{C}$ and $V_D$. Since $D$ has no children in $N$, in line 5 we
assign $w(V_D)$ to $1$. Next, we process the soft constraint between $V_{A,B}$
and $V_C$: $V_D$ is the only child of $V_C$, hence $w(V_C) = w(V_D) \times
\cD(V_D) = 1 \times 2 = 2$. Subsequently, since $V_C$ is the only child of both
$V_A$ and $V_B$, we assign $w(V_A) = w(V_B) = w(V_C) \times |\cD(V_C)|=2 \times
2 =4$.

Now, consider two outcomes $o_1=abcd$ and $o_2=a \bar{b} cd$. The total penalty
of $o_1$ is $(w(V_A) \times p_1) + (w(V_B) \times p_1) + (w(V_C) \times p_1) +
(w(V_D) \times p_1) = 0$, since $p_1=0$, while the total penalty of $o_2$ is
$(w(V_A) \times p_1) + (w(V_B) \times p_2) + (w(V_C) \times p_2) + (w(V_D)
\times p_1) = (4 \times 1)+(2 \times 1) = 6$ since $p_2=1$.  Therefore, we can
conclude that $o_1$ is better than $o_2$ since $min(0,6)=0$.

We now prove that our algorithm for weight computation 
ensures the cp-condition on the resulting 
set of soft constraints, and this also implies preserving
the ordering information with respect to the original CP-net.

\begin{theorem}
\label{preserving}
The SC-net based weighted SCSP $N_c$, generated from an acyclic CP-net $N$,
is an information preserving approximation of $N$, i.e.\ for each pair of
outcomes $\alpha,\beta$ we have $\alpha \succ \beta \;\Rightarrow\; 
\alpha >_{min+} \beta$.
\end{theorem}

\proof{ 
Due to the CP-net semantics, %to prove this claim 
it is enough to show
that, for each variable $X \in N$, each assignment $\bu$ on $Pa(X)$, and 
each pair of values $x_1,x_2 \in \cD(X)$, if CP-net specifies that 
$u : x_1 \succ x_2$, then we have $x_1\bu\by >_{min+} x_2\bu\by$, 
for all assignments $\by$ on $\bY = \bV - \left\{\{X\}\cup Pa(X)\right\}$.
  By definition, $x_1\bu\by >_{min+} x_2\bu\by$ iff
{\scriptsize
%  \[
$  \sum_{s \in S} p'((x_1 uy)_{|_s}) < \sum_{s \in S} p'((x_2 uy)_{|_s}),
$ %  \]
} 
where $S$ is the set of soft constraints of $N_c$ and notation $(x_1 
uy)_{|_s}$stands for the projection on the outcome on constraint $s$. The 
constraints on 
which $x_1 uy$ differs from $x_2uy$ are: 
constraint $c$ on $V_{Pa(X)}$ and $V_X$, and
all the constraints $t_i \in T$ on $V_{Pa(B_i)}$ and $V_{B_i}$ such 
that $X \in Pa(B_i)$ (in what follows, we denote the children 
of $X$ by $\cB=\{V_{B_1}, \cdots, V_{B_h}\}$). Thus, we can 
rewrite the above inequality as 
{\scriptsize
$ %  \[
  p'((x_1 uy)_{|_c}) + \sum_{t_i \in T} p'((x_1 uy)_{|_{t_i}}) <
  p'((x_2 uy)_{|_c}) + \sum_{t_i \in T} p'((x_2 uy)_{|_{t_i}})
%  \]
$  }
By construction of $N_c$ we have
  {\scriptsize
$ %  \[
p'(\pi_c(x_1uy)_{|_c})=w_c \times p(x_1 u) < p'((x_2 uy)_{|_c})=w_c \times p(x_2 
u)
$ %  \]
  }
  and thus $x_1\bu\by >_{min+} x_2\bu\by$ iff
  {\scriptsize
$ %  \[
  w_c p(x_2 u) - w_c p(x_1 u)>
  \sum_{t_i \in T} p'((x_1 uy)_{|_{t_i}}) - \sum_{t_i \in T} 
p'((x_2uy)_{|_{t_i}})
$ %  \]
  }
  In particular, this will hold if
  {\scriptsize
$ %  \[
  w_{c} (min_{x,x' \in \cD(X)}|p(x u)-p(x' u)|) >
  \sum_{t_i \in T} w_{t_i} (max_{x,x',z,b} |p(x' zb)-p(x zb)|)
$ %  \]
} where $z$ is the assignment to all parents of $\cB$ other than $X$.  Observe
,
that the maximum in the right term is obtained 
when $p(x' zb)=|\cD(\cB)|-1$ and $p(x zb)=0$.  On the other hand, 
$min_{x,x' \in \cD(X)}|p(x' u)-p(x u)|=1$.  In other words: 
$ w_{c}> \sum_{t_i \in T} w_{t_i} (|\cD(B_i)|-1)$ must hold.  But
this is ensured by the algorithm, setting (in line 7) $w_c=\sum_{t_i \in T}
w_{t_i} (|\cD(B_i)|$.
}

\begin{theorem}[Complexity]
%carmel-f : rephrased 
%% Given an acyclic CP-net $N$, where the number of parents of a node 
%% is constantly bound, 
  Given an acyclic CP-net $N$ with the node in-degree bounded by a
  constant, the construction of the corresponding SC-net based
  weighted SCSP $N_c$ is polynomial in the size of $N$.
\end{theorem} 

\proof{ 
If the CP-net has $n$ nodes then the number of vertices 
$V$ of the derived SC-net is at most $2n$. 
In fact, in the SC-net a node representing a 
feature appears at most once and there is 
at most one node representing its parents. 
%The worst case is 
%when for each node 
%we must introduce a new node representing its parents; 
%this happens when each feature depends on 
%more than one feature. So to the $n$ features we must 
%add $n$ nodes of type parents \footnote{If 
%there are $I$ independent features we will have to add at most $n-I$ new 
%nodes.}. 
If the number of edges of the CP-net 
is $e$, then the number of edges $E$ in 
the SC-net (including hard and soft edges) is at most $e+n$, 
since each edge in the CP-net corresponds to at most one constraint, 
and each feature in the CP-net 
generates at most one new soft constraints. 
Topological sort can be performed in $O(V+E)$, that is, $O(2n+e+n) = 
O(e+n)$. Then, for each 
node, that is, $O(V)$ times, at most $V$ children must be checked to 
compute the new weight value, leading to a number of 
checks which is $O(V^2) = O(n^2)$. 
Each check involves checking a number of assignments 
which is exponential in the number of parents of a node.
Since we assume that the number of parents of a node is limited
by a constant, this exponential is still a constant. 
Thus the total time complexity is $O(V^2)$ (or $O(n^2)$ if we 
consider the size of the CP-net). 
} 

%\begin{lemma}[worsening flip in min+ model] 
%A worsening flip in the original model is equivalent to substituting a tuple 
%in the relation of one ore more constraints with a tuple with higher penalty. 
%\end{lemma} 
%\proof{Consider $\stackrel{\rightarrow}{v}=v_1v_2\cdots v_n$. To perform a 
%legal worsening flip on $\stackrel{\rightarrow}{v}$ means to select a feature, 
%lets say $F_i$, either that is independent from other features but its 
%value, $v_i$ in $\stackrel{\rightarrow}{v}$ isn't the most preferred among 
%the values in its domain or that is dependent on other features, $F_{i1}, 
%\cdots, F_{iq}$ s.t. $ \forall ij ij\leq i$ and such that 
%$v_i$ can be worsened with respect to the assignments 
%$v_{i1},\cdots ,v_{iq}$. A flip is simply a substitution of $v_i$ with 
%$v_i'$ such that either $v_i \succ v_i'$, independently from other features 
%or $v_{i1}\cdots v_{iq}v_i \succ v_{i1}\cdots v_{iq}v_i'$. 
%The way we have defined the constraint network and the preference function 
%assigning penalties implies that 
%$$ 
%v_i \succ v_i' \Rightarrow f(v_i)\leq f(v_i') 
%$$ 
%on the unary constraint on the variable representing $F_i$, 
%and 
%$$ 
%v_{i1}\cdots v_{iq}v_i \succ v_{i1}\cdots v_{iq}v_i' \Rightarrow 
%f(v_{i1}\cdots v_{iq}v_i)\leq f(v_{i1}\cdots v_{iq}v_i') 
%$$ 
%on the constraint between the variable representing $F_{i1},\cdots , F_{iq}$ 
%and the variable representing $F_i$. 
%} 

Let us compare in more details the original preference relation induced by the
CP-net and this induced by its min+ semi-ring based SC-net.  The comparison is
summarized in the following table, where $\sim$ denotes incomparability. Notice
that Theorem~\ref{preserving} shows that ordering information is preserved by
the approximation.

{\scriptsize
 \begin{center} 
    \begin{tabular}{|c|c||c|c|} 
      \hline 
        \multicolumn{2}{|c||}{{\bf CP-nets} $\Rightarrow$ min+} 
& \multicolumn{2}{c|}{{\bf min+} $\Rightarrow$ CP-nets}\\ 
        \hline 
        \hline 
        $\prec$ & $<$ & $<$ & $\prec, \sim $\\ 
\hline
        $\succ$ & $>$ & $>$ & $\succ, \sim$\\ 
          \hline 
         $\sim$ & $<,>,=$ & $=$ & $=,\sim$\\ 
        \hline 
    \end{tabular} 
  \end{center} 
}

Since the min+ approximation is a total ordering, it 
is a linearization of the original partial ordering. 
In compensation, however, preferential comparison 
is now linear time.

\subsection{SLO soft constraints} 

We also consider a different semi-ring to approximate 
CP-nets via soft constraints. The SLO c-semi-ring is defined as follows: 
$S_{SLO}=\langle A,max_s,min_s,${\bf MAX},{\bf 0}$\rangle$, where 
$A$ is the set of sequences of $n$ integers from 0 to MAX, 
{\bf MAX} is the sequence of $n$ elements all equal to MAX, and 
{\bf 0} is the sequence of $n$ elements all equal to 0. 
The additive operator, $max_s$ and the multiplicative operator, 
$min_s$ are  defined as follows: 
given $s=s_1\cdots s_n$ and $t=t_1\cdots t_n$, 
$s_i=t_i, i=1 \leq k$ and $s_{k+1} \neq t_{k+1}$, 
then $max_s(s,t)= s$ if $s_{k+1} \succ t_{k+1}$ 
else $max_s(s,t) = t$; 
on the contrary, $min_s(s,t)= s$ if $s_{k+1} \prec t_{k+1}$ 
else $min_s(s,t) = t$. 

It is easy to show that $S_{SLO}$ is a c-semi-ring 
% deriving the properties 
%of $max_s$ and $min_s$ from the ones of $max$ and $min$. It is also 
and that the ordering induced by $max_s$ on $A$ is lexicographic ordering
\cite{lex}. To
model a CP-net as a soft constraint problem based on $S_{SLO}$, we set MAX
equal to the cardinality of the largest domain - 1, and $n$ equal to the number
of soft constraints of the SC net. All the weights of the edges are set to 1.
Considering the binary soft constraint 
on $Pa(X)=\{U_1 \dots U_h\}$ and $X$, a tuple of assignments 
$(u_1, \dots, u_h,x)$ will be assigned, as preference,
the sequence of $n$ integers:
$(MAX,MAX,\dots, MAX-i+1, \dots, MAX)$. 
In this sequence, each 
element corresponds to a soft constraint. The element corresponding to
the constraint on $Pa(X)$ and $X$ is $MAX-i+1$,
where $i$ is the distance from the top of the total order of 
the value $x$ (i.e. we have a preference statement of the form
$u: x_1 \succ x_2 \succ \dots x_i=x \succ x_{|D(X)|}$).     
In the example shown in Figure \ref{scnet}, all the preferences will be lists
of four integers (0 and 1), where position $i$ corresponds to constraint with
weight $w_i$. For example, in constraint weighted $w_3$, $p_1=(1,1,1,1)$ and
$p_2=(1,1,0,1)$. Given the pair of outcomes $o_1=abcd$ and $o_2=a \bar{b} cd$,
the global preference associated with $o_1$ is $(1,1,1,1)$, since it does not
violate any constraint, while the preference associated with $o_2$ is
$min_S\{(1,1,1,1),(1,0,1,1),(1,1,0,1),(1,1,1,1)\}=(1,0,1,1)$.  We can conclude
that $o_1$ is better than $o_2$.

Similar to the comparison performed for min+ semi-ring, the following
table compares the preference
relation induced by the SLO semiring and that
induced by the CP-net. 

{\scriptsize
\begin{center} 
    \begin{tabular}{|c|c||c|c|} 
      \hline 
        \multicolumn{2}{|c||}{{\bf CP-nets} $\Rightarrow$ SLO} 
& \multicolumn{2}{c|}{{\bf SLO} $\Rightarrow$ CP-nets}\\ 
%        {\bf CP-nets} & SLO & {\bf SLO} & CP-nets\\ 
        \hline 
        \hline 
        $\prec$ & $<$ & $<$ & $\prec,\sim$\\ 
\hline
        $\succ$ & $>$ & $>$ & $\succ,\sim$\\ 
          \hline 
          = & = & = & =\\ 
        \hline 
        $\sim$ & $<,>$ & \multicolumn{2}{c|}{\mbox{}} \\ % &  \\ 
        \hline 
    \end{tabular} 
  \end{center} 
}

Note that the SLO model both preserves information and ensures the
cp-condition. The proof of this is straightforward and is omitted due
to lack of space.
The SLO model, like the weighted model, is very useful to answer dominance queries as it inherits the
linear complexity from its semi-ring structure. In addition, the sequences of
integers show directly the ``goodness'' of an assignment, i.e. where it
actually satisfies the preference and where it violates it.

\subsection{Comparing the two approximations} 

Given an acyclic CP-net $N$, let $N_c^{min+}$ and $N_c^{SLO}$ stand for the
corresponding min+ and SLO based SC-nets respectively.  From the results in
the previous section, we can see that pairs of outcomes ordered by $N$ remain
ordered the same way by both $N_c^{min+}$ and $N_c^{SLO}$.  On the other
hand, pairs of outcomes incomparable in $N$ are distributed among the three
possibilities (equal or ordered in one the two ways) in $N_c^{min+}$, while
being strictly ordered by $N_c^{SLO}$.  Therefore, the (total) preference
relation induced by $N_c^{min+}$ is a less brutal linearization of the partial
preference relation induced by $N$, compared to that induced by $N_c^{SLO}$.
Mapping incomparability onto equality might seem more reasonable
than mapping it onto an arbitrary strict ordering, since the choice is still
left to the user.  We might conclude %therefore 
that the min+ model is to be
preferred to the SLO model, as far as approximation %of the original model
is concerned. However, maximizing the minimum reward, as in any fuzzy framework
\cite{schiex}, has proved its usefulness in problem representation. The user
may therefore need to balance the linearization of the order and the
suitability of the representation provided.

\section{Future Work} %Conclusions}

%We have proposed a unifying modelling and solving environment 
%in which both hard and soft constraints as well 
%as qualitative conditional preferences can be handled efficiently.
%The framework consists of a soft constraint solver plus 
%n algorithm for approximating the 
%semantics of conditional preference statements by translating
%them into soft constraints. The translation requires some approximation,
%but offers a computational gain.
%We have also studied the complexity of consistency checking 
%for general sets of conditional preference statements.

%In the future, w
We plan to use our approach in a preference elicitation 
system in which we guarantee the consistency of the user preferences, and 
guide the user to a consistent scenario. 
Morover, we also plan to exploit the use of partially ordered 
preferences, as allowed in soft constraints, to better approximate 
CP nets.
%Moreover, we also plan to 
%study the issue of abstracting one order with another one, 
%which has been considered here in several instances. 
%We also plan to study experimentally the phase transition in 
%the satisfiability of conditional 
%preference statements. 
Finally, we intend to use machine learning techniques to 
learn conditional preferences from comparisons 
of complete assignments. 

%\end{document}

%\bibliographystyle{aaai} 
%\bibliographystyle{named-5} 
%\bibliography{softcsp} 

\end{document}